\documentclass{article}
\usepackage{iclr2027_conference,times}

\usepackage{amsmath,amsfonts,bm}

\def\eqref#1{equation~\ref{#1}}

\def\1{\bm{1}}

\DeclareMathAlphabet{\mathsfit}{\encodingdefault}{\sfdefault}{m}{sl}
\SetMathAlphabet{\mathsfit}{bold}{\encodingdefault}{\sfdefault}{bx}{n}

\usepackage{hyperref}
\hypersetup{hidelinks}
\usepackage{url}
\usepackage{booktabs}
\usepackage{amsmath,amssymb}
\usepackage{bm}
\usepackage{multirow}
\usepackage{makecell}
\usepackage{graphicx}
\usepackage{subcaption}
\usepackage{algorithm,algorithmic}
\usepackage{amsthm}

\newtheorem{proposition}{Proposition}

\title{Self-Privileged Fine-Tuning}

\author{
  Yangrong Hu\textsuperscript{1,*},
  Xueyu Zhou\textsuperscript{1,*},
  Yue Wu\textsuperscript{1},
  Jian Huang\textsuperscript{1,2,\textdagger} \\
  \small\textsuperscript{1}Department of Data Science and Artificial Intelligence \\
  \small\textsuperscript{2}Department of Applied Mathematics \\
  \small The Hong Kong Polytechnic University \\
  \small\textit{\{yangrong.hu,xueyu.zhou,yue0301.wu\}@connect.polyu.hk} \\
  \small\textit{j.huang@polyu.edu.hk}
}

\begin{document}

\iclrfinalcopy
\maketitle
\pagestyle{empty}
\lhead{}
\begingroup
\renewcommand{\thefootnote}{*}
\footnotetext{Equal contribution.}
\endgroup
\begingroup
\renewcommand{\thefootnote}{\textdagger}
\footnotetext{Corresponding author: Jian Huang.}
\endgroup

\begin{abstract}
Supervised fine-tuning (SFT) is a key component of language model post-training, enabling effective task adaptation and serving as an important foundation for subsequent reinforcement learning. Recent methods seek to improve SFT by introducing token-level weights into the training objective, where these weights are typically derived from a model’s output distribution at each token position. 
To better understand how such weighting schemes affect SFT, we analyze their optimization objectives and theoretically characterize the target distributions they induce. Our analysis shows that dynamic weighting methods, such as Dynamic Fine-Tuning (DFT), induce a heavily sharpened distribution, whereas static weighting methods fit a target distribution proportional to the product of the data distribution and a frozen prior.
These findings suggest that static weighting provides a more stable training target, while its effectiveness critically depends on the quality of the prior distribution.
Building on this observation, we propose Self-Privileged Fine-Tuning (SPFT), which uses a frozen copy of the initial model conditioned on privileged information (such as ground truth for mathematical questions) as a prior to provide guidance for token-level weighting during fine-tuning. 
Through a Bayesian decomposition, we further show that SPFT reweights the standard static prior according to how informative each target token is for predicting the ground-truth outcome, thereby assigning larger weights to more predictive tokens.
Experiments on mathematical reasoning and code generation show that SPFT improves over standard SFT and other baselines across the main settings.
\end{abstract}
\section{Introduction}
\label{sec:intro}

Supervised fine-tuning (SFT) is a fundamental component of language-model post-training and commonly serves as the foundation for subsequent reinforcement learning (RL)~\citep{ouyang2022training, touvron2023llama}.
Whereas RL learns from reward signals assigned to model-generated responses, SFT directly learns from supervised datasets by maximizing the likelihood of target responses, which enables pretrained models to better adapt to downstream tasks such as instruction following, mathematical reasoning and code generation.

As standard SFT assigns equal weight to all target tokens, recent studies have explored improving SFT by introducing non-uniform token weights.
\citet{wu2025generalization} reweight each token according to its probability under the current model, giving larger updates to tokens that are better supported by the model. Building on this formulation, \citet{zhu2026anchored} introduce an anchoring term based on KL divergence to stabilize optimization, while \citet{lin2026sft} introduce temperature scaling to control the magnitude of token weights. Despite their different designs, these methods derive weights from the model being optimized, causing the weighting signal to evolve dynamically throughout training. More recently, \citet{wang2026prift} take a different approach by computing token weights from a frozen reference model, resulting in a static weighting scheme. While these methods demonstrate the effectiveness of token reweighting, the mechanisms underlying their improvements remain insufficiently understood.

To better understand these methods, we analyze their loss functions and induced optimization objectives in Section~\ref{sec:theory}. For dynamic weighting methods such as Dynamic Fine-Tuning (DFT)~\citep{wu2025generalization}, the target distribution evolves with the model throughout training, leading to progressive sharpening and increasing concentration on high-probability modes. In contrast, static weighting such as PriFT~\citep{wang2026prift} induces a fixed target distribution proportional to the product of the data distribution and the prior distribution. This characterization makes the quality of the prior central to static token weighting.

Motivated by this observation, we ask whether the model itself can provide a more informative prior for static weighting without relying on an external teacher.
To this end, we propose \textbf{Self-Privileged Fine-Tuning (SPFT)}. 
SPFT uses a frozen copy of the initial model conditioned on privileged information as a prior to guide token-level weighting during fine-tuning. 
A Bayesian decomposition shows that the resulting privileged prior modifies the standard static prior according to the predictive contribution of each target token to the ground-truth outcome, so that more informative tokens receive larger weights.
Experiments on mathematical reasoning and code generation show that SPFT consistently outperforms standard SFT and other token-weighting baselines across the main settings.

Our contributions are:
\begin{itemize}
    \item We provide a theoretical characterization of token-weighted supervised fine-tuning, showing that dynamic weighting progressively sharpens the model distribution, while static weighting optimizes toward a fixed distribution jointly determined by the data distribution and a frozen prior.
    \item We propose \textbf{Self-Privileged Fine-Tuning (SPFT)}, which constructs a more informative static prior by conditioning a frozen copy of the initial model on privileged information available during training. A Bayesian decomposition further shows that SPFT upweights tokens according to their contribution to predicting the ground-truth outcome.
    \item We evaluate SPFT on mathematical reasoning and code generation, where it consistently improves over standard SFT and existing token-weighting baselines. Further ablation studies validate the importance of a fixed privileged prior and examine other key design choices in SPFT.
\end{itemize}

\section{Related Work}
\label{sec:related_work}

\paragraph{Token-Reweighted Supervised Fine-Tuning.}
SFT remains a stable and data-efficient post-training paradigm, but its generalization can lag behind reinforcement learning in some settings~\citep{chu2025sft, shenfeld2025rl}. RL methods such as PPO and verifiable-reward reasoning training optimize the model through on-policy rollouts and rewards, yet online sampling introduces substantial computational cost~\citep{schulman2017proximal, shao2024deepseekmath, guo2025deepseek, sheng2024hybridflow, swamy2025roads}. In contrast, offline preference optimization methods such as DPO avoid explicit online rollouts~\citep{rafailov2023direct}. Recent studies introduce token-level reweighting as an offline bridge between SFT and RL~\citep{wu2025generalization, zhu2026anchored}. Existing methods can be broadly divided into online dynamic weighting and anchored or prior-guided static weighting. DFT~\citep{wu2025generalization}, temperature-based variants~\citep{lin2026sft}, and related uncertainty or confidence-based schemes use statistics from the evolving model during training, which can improve performance on difficult reasoning examples. Anchored SFT~\citep{zhu2026anchored} and PriFT~\citep{wang2026prift} improve stability through reference anchoring or a frozen pretrained prior, but their signals remain unconditioned on the task outcome. SPFT instead conditions the frozen reference model on the verified outcome, producing a goal-directed process signal while remaining offline and independent of the evolving model policy.

\paragraph{Privileged Information and Knowledge Distillation.}
Knowledge distillation traditionally matches a teacher's soft predictions on fixed training prefixes~\citep{hinton2015distilling}. Autoregressive methods such as MiniLLM~\citep{gu2024minillm} and Generalized Knowledge Distillation (GKD)~\citep{agarwal2024onpolicy} address the mismatch between fixed prefixes and student-generated trajectories through on-policy sampling, but require online rollouts and, in GKD, an external teacher. On-Policy Self-Distillation (OPSD)~\citep{zhao2026selfdistilledreasoneronpolicyselfdistillation} removes the external-teacher requirement by conditioning on a reference solution as privileged information. SPFT adopts this idea offline, using verified outcomes to construct token weights for SFT without on-policy sampling or full-vocabulary distribution matching.

\section{Theoretical Analysis of Token-Weighted Fine-Tuning}
\label{sec:theory}

In this section, we analyze the optimization behavior induced by different token-weighting strategies. We first examine dynamic weighting and characterize how its induced target distribution evolves during training. We then analyze static weighting and show that it instead defines a fixed target distribution determined by the data distribution and the prior.

\paragraph{ Notation}

Let $x$ denote the prompt and $y=(y_1,y_2,...,y_T)$ denote the corresponding output. We denote the underlying data distribution by $p_{\mathrm{data}}(x,y)$. Let $p_{\theta_0}(y\mid x)$ denote the initial model's distribution and $p_{\theta_k}(y\mid x)$ the distribution at optimization step $k$.

\subsection{Dynamic Weighting Leads to Progressive Distribution Sharpening}
\label{sec:dynamic_weighting}

We first analyze dynamic weighting through DFT~\citep{wu2025generalization}, where the training weight is determined by the current model and therefore evolves throughout training.
The objective of DFT is given by
\begin{equation}
	\mathcal{L}_{\mathrm{DFT}}(\theta)
	=
	-
	\mathbb{E}_{(x,y)\sim p_{\mathrm{data}}}
	\left[
	\sum_{t}
	\operatorname{sg}\!\left[
	p_{\theta}(y_t\mid x,y_{<t})
	\right]
	\log p_{\theta}(y_t\mid x,y_{<t})
	\right],
	\label{eq:dft_objective}
\end{equation}
where $\operatorname{sg}[\cdot]$ denotes the stop-gradient operator.

At the $k$-th optimization step, the weighting term is evaluated using the current model $p_{\theta_k}$ and treated as fixed during this step.
Accordingly, the objective at the $k$-th step can be written as
\begin{equation}
	\mathcal{L}_{\mathrm{DFT}}^{(k)}(\theta)
	=
	-
	\mathbb{E}_{(x,y)\sim p_{\mathrm{data}}}
	\left[
	\sum_{t}
	\operatorname{sg}\!\left[p_{\theta_k}(y_t\mid x,y_{<t})\right]
	\log p_{\theta}(y_t\mid x,y_{<t})
	\right].
	\label{eq:dft_step_objective}
\end{equation}

For a given token context $(x,y_{<t})$, the optimal solution to the objective in Eq.~\ref{eq:dft_step_objective} satisfies
\begin{equation}
	p_{\theta_{k+1}}(y_t\mid x,y_{<t})
	\propto
	p_{\mathrm{data}}(y_t\mid x,y_{<t})
	p_{\theta_k}(y_t\mid x,y_{<t}).
	\label{eq:dft_step_optimum}
\end{equation}
As optimization proceeds, the target distribution at each token context depends on the model obtained from the preceding step, causing the optimization target to evolve continuously throughout training.
When each optimization step reaches the corresponding optimum, repeatedly applying Eq.~\ref{eq:dft_step_optimum} gives
\begin{equation}
	p_{\theta_{k+1}}(y_t\mid x,y_{<t})
	\propto
	p_{\theta_0}(y_t\mid x,y_{<t})
	p_{\mathrm{data}}(y_t\mid x,y_{<t})^{k+1}.
	\label{eq:dft_closed_form}
\end{equation}

For two candidate tokens $v_1$ and $v_2$ under the same context $(x,y_{<t})$, the probability ratio satisfies
\begin{equation}
	\frac{
		p_{\theta_{k+1}}(v_1\mid x,y_{<t})
	}{
		p_{\theta_{k+1}}(v_2\mid x,y_{<t})
	}
	=
	\frac{
		p_{\theta_0}(v_1\mid x,y_{<t})
	}{
		p_{\theta_0}(v_2\mid x,y_{<t})
	}
	\left(
	\frac{
		p_{\mathrm{data}}(v_1\mid x,y_{<t})
	}{
		p_{\mathrm{data}}(v_2\mid x,y_{<t})
	}
	\right)^{k+1}.
	\label{eq:dft_probability_ratio}
\end{equation}
Therefore, whenever
$p_{\mathrm{data}}(v_1\mid x,y_{<t})
>
p_{\mathrm{data}}(v_2\mid x,y_{<t})$,
the relative probability assigned to $v_1$ grows exponentially with the number of optimization steps.
As a result, tokens with higher conditional probability under the data distribution are progressively amplified relative to lower-probability tokens, causing the next-token distribution to become increasingly sharp throughout training.

\begin{proposition}[Limiting Distribution of DFT]
	\label{prop:dft_limit}

	Assume that the maximizer is unique, and let
	$v^{*}=\arg\max_{v} p_{\mathrm{data}}(v\mid x,y_{<t})$.
	Also assume that the initial model assigns nonzero probability to this mode, i.e.,
	$p_{\theta_0}(v^{*}\mid x,y_{<t})>0$.
	Then,
	\begin{equation}
		\lim_{k\to\infty}
		p_{\theta_k}(v\mid x,y_{<t})
		=
		\mathbf{1}[v=v^{*}].
		\label{eq:dft_limit}
	\end{equation}
\end{proposition}

Proposition~\ref{prop:dft_limit} shows that DFT progressively sharpens the next-token distribution and eventually concentrates all probability mass on the unique mode of the data distribution.
This analysis also suggests that similar sharpening behavior may arise in other dynamic weighting methods whose evolving weights progressively emphasize high-probability regions of the data distribution.

\begin{figure*}[t]
\centering
\begin{subfigure}[t]{0.24\textwidth}
  \includegraphics[width=\linewidth]{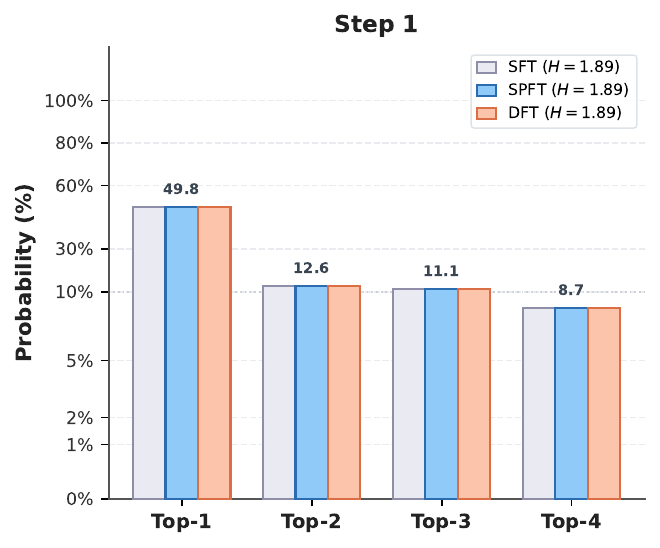}
\end{subfigure}\hfill
\begin{subfigure}[t]{0.24\textwidth}
  \includegraphics[width=\linewidth]{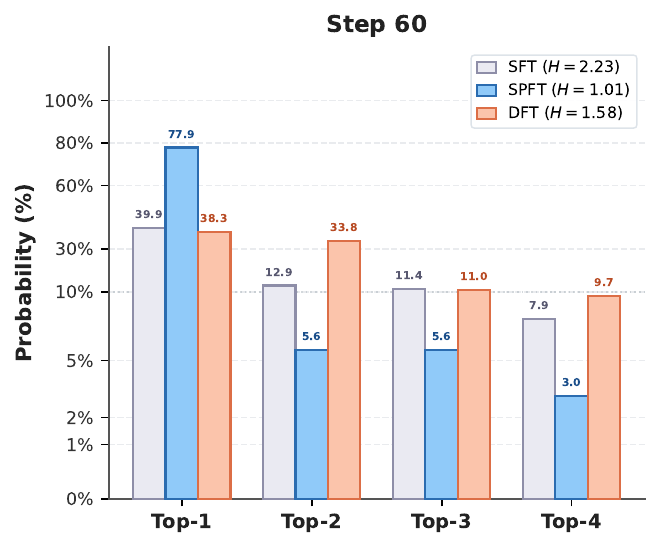}
\end{subfigure}\hfill
\begin{subfigure}[t]{0.24\textwidth}
  \includegraphics[width=\linewidth]{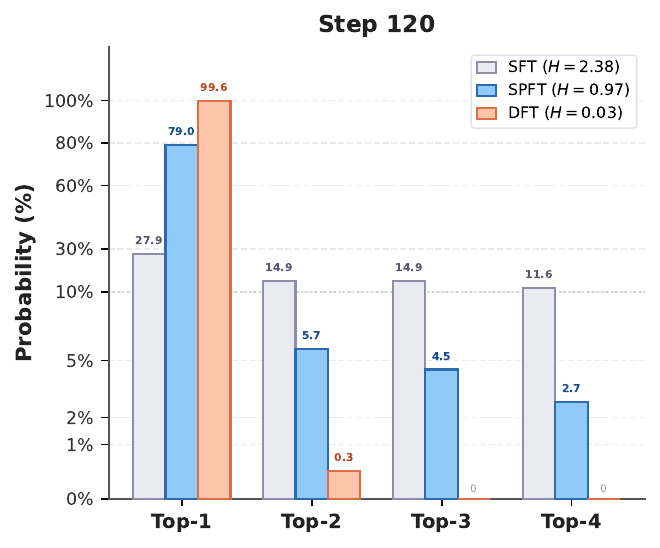}
\end{subfigure}\hfill
\begin{subfigure}[t]{0.24\textwidth}
  \includegraphics[width=\linewidth]{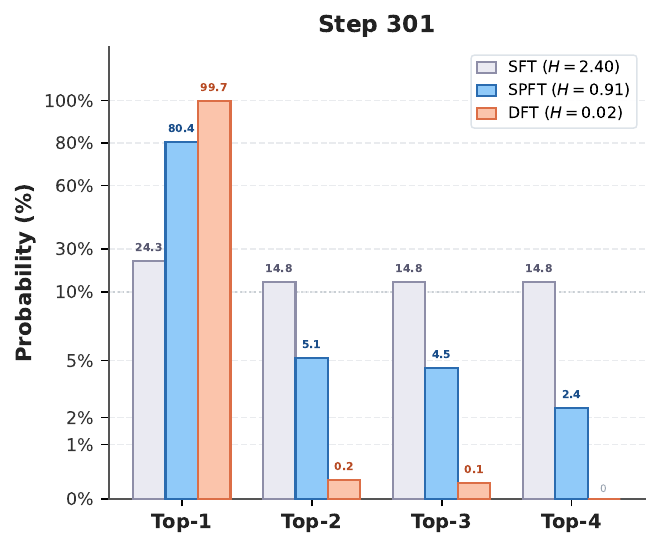}
\end{subfigure}
\caption{Next-token distribution evolution of SFT, DFT, and SPFT at four training steps.}
\label{fig:probe_distribution_sharpening}
\end{figure*}

Consistent with this analysis, Figure~\ref{fig:probe_distribution_sharpening} tracks the next-token distribution at four training steps and provides an illustration of the sharpening behavior. While SFT remains comparatively diffuse, DFT rapidly concentrates nearly all probability mass on the top-ranked token, whereas SPFT further sharpens the distribution relative to SFT, but in a selective manner: tokens receiving stronger support from the privileged prior are amplified more substantially, while alternative tokens retain non-trivial probability mass.

\subsection{Static Weighting Induces a Fixed Product Distribution}
\label{sec:static_weighting}

We next consider static weighting methods, where the token weights are determined by a fixed reference model and remain unchanged throughout training.
Let $p_{\mathrm{prior}}$ denote the frozen reference distribution.
The corresponding weighted fine-tuning objective can be written as
\begin{equation}
	\mathcal{L}_{\mathrm{static}}(\theta)
	=
	-
	\mathbb{E}_{(x,y)\sim p_{\mathrm{data}}}
	\left[
	\sum_t
	p_{\mathrm{prior}}(y_t\mid x,y_{<t})
	\log p_{\theta}(y_t\mid x,y_{<t})
	\right].
	\label{eq:static_objective}
\end{equation}

\begin{proposition}[Optimal Distribution of Static Weighting]
	\label{prop:static_optimum}
	For a fixed token context $(x,y_{<t})$, the optimal solution to the objective in Eq.~\ref{eq:static_objective} satisfies
	\begin{equation}
		p_{\theta}^{*}(v\mid x,y_{<t})
		\propto
		p_{\mathrm{data}}(v\mid x,y_{<t})
		p_{\mathrm{prior}}(v\mid x,y_{<t}).
		\label{eq:static_optimum}
	\end{equation}
\end{proposition}

Proposition~\ref{prop:static_optimum} shows that static weighting defines a fixed target distribution determined jointly by the data distribution and the reference distribution.
In particular, for any two candidate tokens $v_1$ and $v_2$ under the same context,
\begin{equation}
	\frac{
		p_{\theta}^{*}(v_1\mid x,y_{<t})
	}{
		p_{\theta}^{*}(v_2\mid x,y_{<t})
	}
	=
	\frac{
		p_{\mathrm{data}}(v_1\mid x,y_{<t})
	}{
		p_{\mathrm{data}}(v_2\mid x,y_{<t})
	}
	\frac{
		p_{\mathrm{prior}}(v_1\mid x,y_{<t})
	}{
		p_{\mathrm{prior}}(v_2\mid x,y_{<t})
	}.
	\label{eq:static_probability_ratio}
\end{equation}
Thus, the relative preference between tokens is determined by both their support under the data distribution and their probability under the prior distribution.
Since $p_{\mathrm{prior}}$ remains unchanged during training, the induced target distribution is fixed throughout optimization, avoiding the continuously evolving target induced by dynamic weighting.

This characterization makes the quality of the prior distribution a central design choice for static weighting.
An informative prior can steer optimization toward more useful target tokens, whereas an inaccurate prior may distort the induced target distribution.
For instance, PriFT~\citep{wang2026prift} uses the frozen initial model as the prior, yielding a stable but outcome-agnostic weighting signal.
This observation motivates the design of a more informative prior that incorporates additional task-relevant information while remaining fixed throughout training.

\providecommand{\Require}{\REQUIRE}
\providecommand{\Ensure}{\ENSURE}
\providecommand{\State}{\STATE}
\providecommand{\While}[1]{\WHILE{#1}}
\providecommand{\EndWhile}{\ENDWHILE}
\providecommand{\ForAll}[1]{\FORALL{#1}}
\providecommand{\EndFor}{\ENDFOR}

\section{Self-Privileged Fine-Tuning}
\label{sec:method}

Building on the theoretical analysis in Section~\ref{sec:theory}, we present the formulation of \textbf{Self-Privileged Fine-Tuning (SPFT)}.
We first describe how privileged conditioning provides an informative reference prior and derive the token-weighted training objective.
We then present a detailed implementation of SPFT in Section~\ref{sec:method_prompts} and Section~\ref{sec:method_sr}.

\subsection{Method Formulation and Objective Function}
\label{sec:method_formulation}

As established in Proposition~\ref{prop:static_optimum}, static token-weighted fine-tuning induces a target distribution proportional to the product of the data distribution and a fixed prior:
\begin{equation}
    p_\theta^*(v \mid x, y_{<t})
    \propto
    p_{\mathrm{data}}(v \mid x, y_{<t})
    \, p_{\mathrm{prior}}(v \mid x, y_{<t}).
    \label{eq:optimal_product_dist}
\end{equation}
Prior work such as PriFT~\citep{wang2026prift} instantiates this prior using the frozen initial model,
$p_{\mathrm{prior}}(y_t \mid x, y_{<t})=p_{\theta_0}(y_t \mid x, y_{<t}).$
While this provides a stable and readily available prior, it may provide limited information about how strongly an intermediate token $y_t$ contributes to the eventual ground-truth outcome, since it is conditioned only on the problem prompt $x$ and the preceding tokens $y_{<t}$.

Drawing inspiration from On-Policy Self-Distillation (OPSD)~\citep{zhao2026selfdistilledreasoneronpolicyselfdistillation}, we incorporate privileged information available during training into the conditioning context of the frozen initial model. Given privileged information $c$ associated with the ground-truth outcome, this yields a prior conditioned on privileged information:
\begin{equation}
    q_t \triangleq p_{\theta_0}(y_t \mid x, c, y_{<t}).
\end{equation}

\begin{proposition}[Privileged Prior as Outcome-Reweighted Prior]
\label{prop:spft_ratio}
Assume that the relevant conditional probabilities are nonzero and let $q_t=p_{\theta_0}(y_t\mid x,c,y_{<t})$
denote the prior conditioned on privileged information $c$.
Then, by Bayes' rule,
\begin{equation}
q_t
=
\frac{
p_{\theta_0}(c\mid x,y_{\le t})
}{
p_{\theta_0}(c\mid x,y_{<t})
}
p_{\theta_0}(y_t \mid x, y_{<t}).
\label{eq:spft_token_ratio}
\end{equation}
\end{proposition}

Proposition~\ref{prop:spft_ratio} shows that privileged conditioning refines the base-model prior by multiplying it with an outcome-likelihood ratio. 
Conditioned on the preceding prefix, this ratio measures the incremental contribution of the current token $y_t$ to the ground-truth outcome $c$. 
A ratio greater than one indicates that $y_t$ contributes positively to deriving the ground-truth outcome, thereby increasing its privileged weight, whereas a ratio below one indicates a weaker contribution toward the outcome and thus reduces its weight.

Using the privileged prior $q_t$ as the token weight, SPFT optimizes the following objective
\begin{equation}
    \mathcal{L}_{\mathrm{SPFT}}(\theta)
    =
    -
    \mathbb{E}_{(x,y,c)\sim p_{\mathrm{data}}}
    \left[
    \sum_t
    q_t
    \log p_\theta(y_t \mid x,y_{<t})
    \right].
    \label{eq:spft_objective}
\end{equation}

The privileged information $c$ is used only to determine the token weights $q_t$, while the model being optimized remains conditioned on the original prompt $x$ and prefix $y_{<t}$. Thus, privileged information affects training through the weighting signal rather than through the model's forward input.

Following Proposition~\ref{prop:static_optimum}, the optimal distribution induced by SPFT satisfies
\begin{equation}
    p_{\theta,\mathrm{SPFT}}^{*}(v\mid x,y_{<t})
    \propto
    p_{\mathrm{data}}(v\mid x,y_{<t})
    \, p_{\theta_0}(v\mid x,c,y_{<t}).
    \label{eq:spft_optimum}
\end{equation}

Using the Bayesian decomposition, this target can be further written as
\begin{equation}
    p_{\theta,\mathrm{SPFT}}^{*}(v\mid x,y_{<t})
    \propto
    p_{\mathrm{data}}(v\mid x,y_{<t})
    \, p_{\theta_0}(v\mid x,y_{<t})
    \frac{
        p_{\theta_0}(c\mid x,y_{<t},v)
    }{
        p_{\theta_0}(c\mid x,y_{<t})
    }.
    \label{eq:spft_optimum_decomposed}
\end{equation}

Compared with PriFT, SPFT therefore retains the original base-model prior while further modulating it according to the contribution of each candidate token to the privileged outcome. Tokens that make the ground-truth outcome more likely receive additional emphasis, whereas tokens that contribute less receive relatively smaller weights. In this way, SPFT preserves the fixed-target property of static weighting while making the induced target distribution more outcome-aware.

\begin{figure*}[t]
\centering
\begin{subfigure}[t]{0.32\textwidth}
  \includegraphics[width=\linewidth]{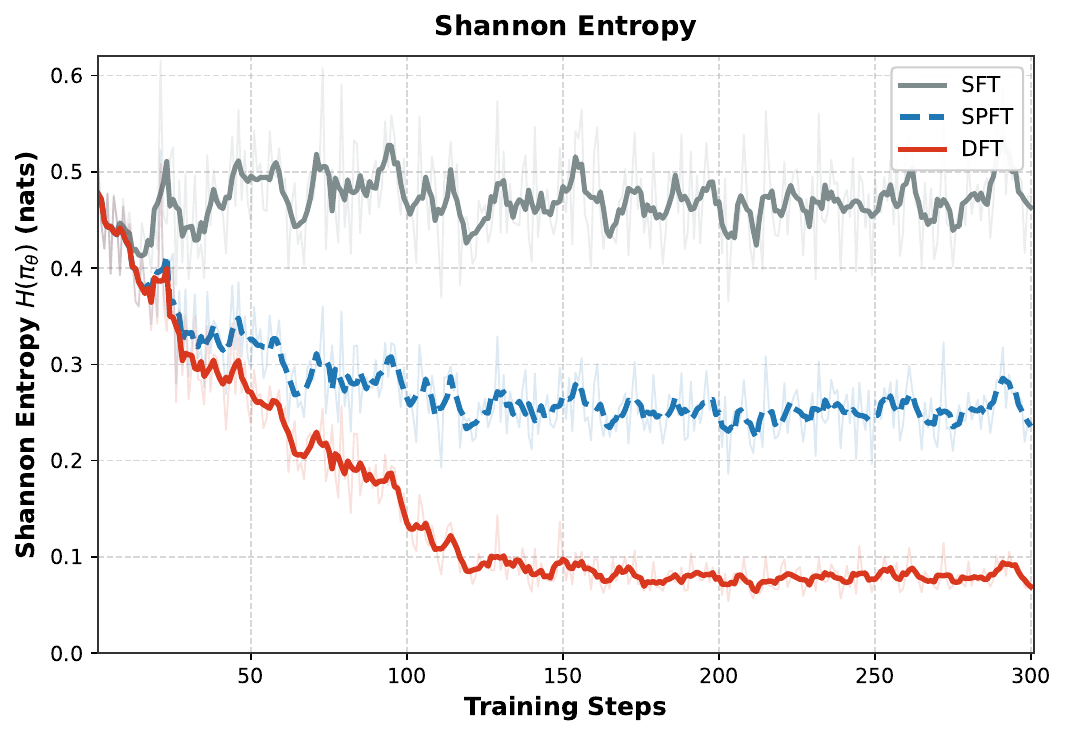}
\end{subfigure}\hfill
\begin{subfigure}[t]{0.32\textwidth}
  \includegraphics[width=\linewidth]{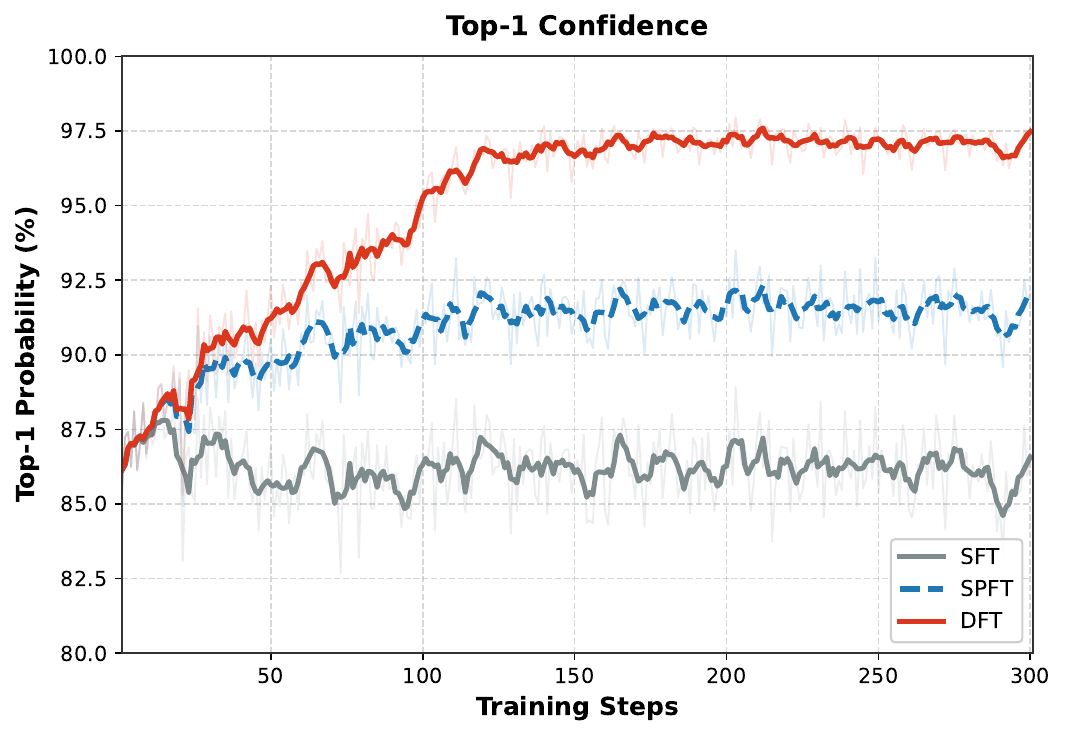}
\end{subfigure}\hfill
\begin{subfigure}[t]{0.32\textwidth}
  \includegraphics[width=\linewidth]{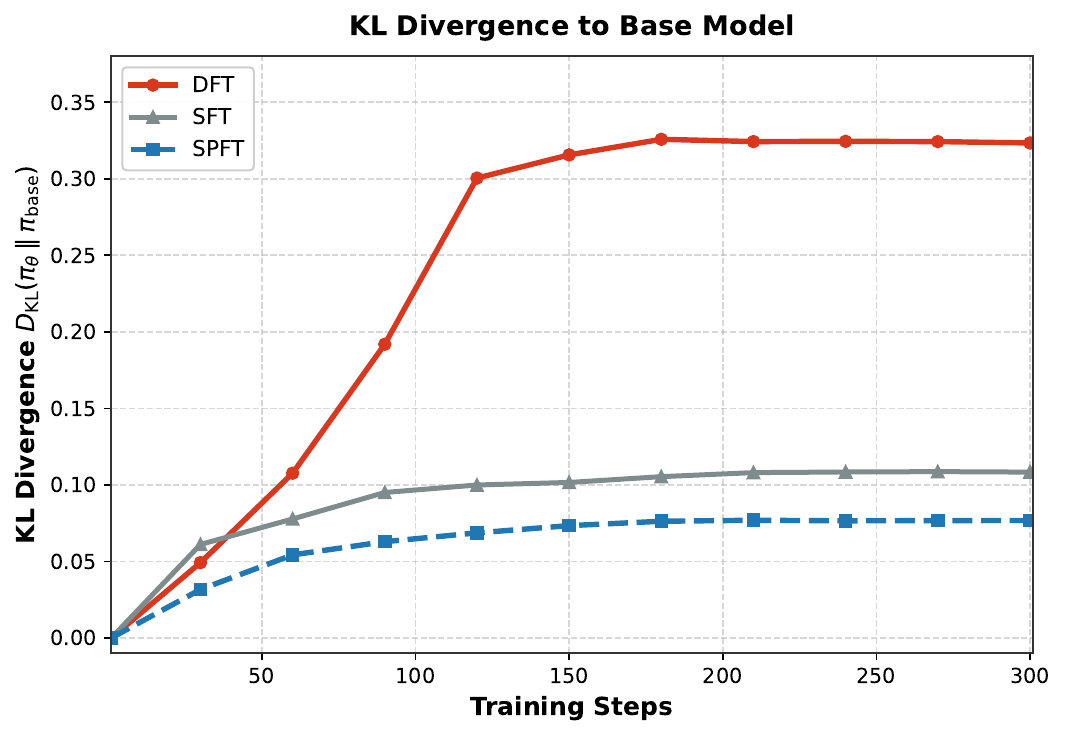}
\end{subfigure}
\caption{
Distributional evolution of SFT, DFT, and SPFT during training, measured by token-level entropy, top-1 confidence, and KL divergence from the initial model. DFT rapidly sharpens the output distribution and moves farthest from the base model, while SPFT produces a more controlled increase in confidence and reduction in entropy with substantially smaller distributional drift.
}
\label{fig:weighting_dynamics_metrics}
\end{figure*}

To quantify how different weighting schemes affect the learned policy distribution, Figure~\ref{fig:weighting_dynamics_metrics} tracks top-1 confidence, token-level entropy, and KL divergence from the initial base model during training.
DFT rapidly increases top-1 confidence while sharply reducing entropy, indicating strong concentration on a narrow set of continuations. Its substantially larger KL divergence further shows that the resulting policy moves farther away from the base model distribution.
By contrast, SFT maintains a relatively flat token distribution, with higher entropy and lower top-1 confidence throughout training. This suggests that uniform token weighting provides limited selective emphasis among candidate continuations and induces comparatively mild distributional change.
SPFT exhibits an intermediate behavior: it increases confidence and reduces entropy relative to SFT, indicating stronger emphasis on selected tokens, while maintaining a much smaller degree of concentration and distributional drift than DFT. 
This is consistent with SPFT selectively amplifying tokens supported by the privileged prior rather than uniformly driving the model toward an increasingly sharp distribution.

\subsection{Privileged Prompt Construction Across Domains}
\label{sec:method_prompts}

Given a training instance $(x, y, c)$, the trainable model receives only the standard task prompt and prefix $h_t = (x, y_{<t})$, whereas the frozen reference model evaluates the sequence under the privileged context $h_t^{\mathrm{P}} = (x, c, y_{<t})$, incorporating the privileged information $c$ within a domain-specific instruction template.

The form of privileged information $c$ differs across reasoning domains.
In mathematical problem solving, $c$ is intuitively instantiated as the ground-truth numerical answer extracted from boxed solutions, guiding the reference model to assess whether intermediate derivations align with the target result.
In program synthesis, however, code tasks lack an obvious concise numerical answer. Simply providing the entire code reference implementation as $c$ causes severe code token leakage.
To provide semantic guidance without leaking surface syntax, we extract high-level conceptual guidance via lightweight static Abstract Syntax Tree (AST) parsing and pattern matching, which  identifies overarching algorithmic paradigms such as dynamic programming, breadth-first search, or two pointers, detects key abstract data structures such as heaps, deques, or state arrays, and isolates boundary invariants such as early return conditions.
This clean-rule abstraction conveys algorithmic intent without exposing implementation details, enabling the privileged prior to reward critical algorithmic choices while maintaining generalization.
The full conditioning interfaces are provided in Appendix~\ref{app:prompts}.

\subsection{Sequence-Relative Weighting (SPFT-SR)}
\label{sec:method_sr}

Absolute privileged likelihoods $q_t$ can vary with overall problem difficulty, obscuring which tokens are relatively well supported within a response. SPFT-SR therefore centers the privileged prior probabilities $q_t$ over the $T$ valid response tokens and retains only positive deviations:
\begin{equation}
    \mu=\frac{1}{T}\sum_{t=1}^{T}q_{t},
    \qquad
    \sigma=\sqrt{\frac{1}{T}\sum_{t=1}^{T}(q_{t}-\mu)^2+\epsilon},
    \qquad
    r_{t}=\left[\frac{q_{t}-\mu}{\sigma}\right]_+.
    \label{eq:spft_sr_score}
\end{equation}
The score $r_t$ selects tokens with above-average likelihood under the privileged prior within the response: tokens with $q_t\leq\mu$ have zero score. This reflects relative outcome-conditioned support rather than guaranteed semantic importance. More details are provided in Appendix~\ref{app:sr_implementation}.
Before the per-response rescaling described in Appendix~\ref{app:sr_implementation}, the score-weighted objective has the form
\begin{equation}
    \mathcal{L}_{\mathrm{SPFT\text{-}SR}}(\theta)
    =
    -
    \mathbb{E}_{(x,y,c)\sim p_{\mathrm{data}}}
    \left[
    \sum_t
    r_t
    \log p_\theta(y_t \mid x,y_{<t})
    \right].
    \label{eq:spft_sr_objective}
\end{equation}

\section{Experiments}
\label{sec:experiments}

\subsection{Experimental Setup}
\label{sec:exp_setup}

\paragraph{Mathematical reasoning.}
We train on a 10K random subsampled dataset from NuminaMath-CoT ~\citep{numina_math_datasets}. We use a global batch size of 32, a learning rate of $10^{-5}$ with cosine decay and 5\% warmup, one epoch, and a maximum sequence length of 3,072 tokens. We evaluate Qwen2.5-Math-1.5B and Qwen2.5-Math-7B~\citep{qwen2024mathtechnicalreport}, together with Qwen3-8B-Base~\citep{qwen2025qwen3technicalreport}. The evaluation uses the five-benchmark suite consisting of MATH-500~\citep{hendrycks2021measuring}, Minerva Math~\citep{lewkowycz2022solving}, OlympiadBench~\citep{he-etal-2024-olympiadbench}, AIME 2024~\citep{aime2024dataset}, and AMC 2023~\citep{amc2023dataset}, with eight samples per problem, temperature $0.7$, top-$p$ $0.8$, evaluation seed 0, and the unified \texttt{qwen-boxed} protocol. For Qwen3-8B, the trained-model comparison uses the strictly matched non-thinking evaluation protocol.

\paragraph{Code generation.}
For code, we train on a 10K randomly subsampled dataset from OpenCodeInstruct~\citep{ahmad2025opencodeinstruct} with a global batch size of 16, a learning rate of $5\times 10^{-5}$, one epoch, and a maximum sequence length of 2,048 tokens. We evaluate Qwen2.5-Coder-1.5B~\citep{hui2024qwen2} using EvalPlus~\citep{evalplus} with greedy decoding. The reported metrics include the original HumanEval~\citep{chen2021evaluating} and MBPP~\citep{austin2021program} benchmarks, together with their EvalPlus-extended HumanEval+ and MBPP+ variants~\citep{evalplus}, and their four-task average.

\paragraph{Baselines.}
We compare SPFT with the pretrained model (Base), standard supervised fine-tuning (SFT)~\citep{chung2022scaling}, dynamic token reweighting (DFT)~\citep{wu2025generalization}, anchored SFT (ASFT)~\citep{zhu2026anchored}, temperature-based reweighting (TALR)~\citep{lin2026sft}, and frozen-prior weighting (PriFT)~\citep{wang2026prift}. All trained methods use the same data, initialization, optimization budget, and evaluation protocol.

\subsection{Experimental Results}
\label{sec:exp_math}
\begin{table*}[t]
\centering
\small
\setlength{\tabcolsep}{4.0pt}
\renewcommand{\arraystretch}{1.12}
\caption{Mathematical reasoning results across five benchmarks. Each benchmark reports Avg@8 / Pass@8, and the final pair gives the unweighted five-benchmark average. External baselines include ASFT~\citep{zhu2026anchored}, TALR~\citep{lin2026sft}, and PriFT~\citep{wang2026prift} whenever a matched result is available.}
\label{tab:math_main}
\resizebox{\textwidth}{!}{\begin{tabular}{>{\centering\arraybackslash}p{1.15cm}lcccccccccccc}
\toprule
\multirow{2}{*}{\textbf{Model}} & \multirow{2}{*}{\textbf{Method}}
& \multicolumn{2}{c}{\textbf{MATH-500}}
& \multicolumn{2}{c}{\textbf{Minerva Math}}
& \multicolumn{2}{c}{\textbf{OlympiadBench}}
& \multicolumn{2}{c}{\textbf{AIME24}}
& \multicolumn{2}{c}{\textbf{AMC23}}
& \multicolumn{2}{c}{\textbf{Average}} \\
\cmidrule(lr){3-4}\cmidrule(lr){5-6}\cmidrule(lr){7-8}\cmidrule(lr){9-10}\cmidrule(lr){11-12}\cmidrule(lr){13-14}
& & Avg@8 & Pass@8 & Avg@8 & Pass@8 & Avg@8 & Pass@8 & Avg@8 & Pass@8 & Avg@8 & Pass@8 & Avg@8 & Pass@8 \\
\midrule
\multirow{8}{*}{\makecell[c]{Qwen2.5\\Math\\(7B)}}
& Base & 60.38 & 87.40 & 16.18 & 42.28 & 27.76 & 54.07 & 14.58 & 36.67 & 49.69 & 85.00 & 33.72 & 61.08 \\
& SFT & 67.85 & 89.40 & 24.45 & 47.43 & 30.96 & 55.85 & 5.42 & 20.00 & 40.62 & 75.00 & 33.86 & 57.54 \\
& DFT & 68.32 & 81.20 & 25.46 & 35.66 & 33.04 & 48.00 & 6.25 & 16.67 & 46.56 & 67.50 & 35.93 & 49.81 \\
& ASFT & 70.32 & 88.00 & 22.75 & 38.24 & 34.50 & 57.04 & 7.08 & 16.67 & 47.81 & 80.00 & 36.49 & 55.99 \\
& TALR & 65.88 & 85.20 & 22.56 & 44.85 & 29.20 & 53.48 & 5.83 & 16.67 & 35.62 & 75.00 & 31.82 & 55.04 \\
& PriFT & 72.08 & 89.20 & 19.49 & 40.07 & 31.93 & 55.56 & 14.58 & 36.67 & 55.31 & 77.50 & 38.68 & 59.80 \\
& SPFT & \textbf{74.05} & 89.40 & 26.61 & 47.43 & \textbf{37.48} & 59.11 & \textbf{15.83} & 40.00 & 59.69 & \textbf{87.50} & \textbf{42.73} & \textbf{64.69} \\
& \textbf{SPFT-SR} & 72.05 & 87.40 & \textbf{30.15} & \textbf{50.00} & 36.89 & 56.44 & 14.58 & 36.67 & 57.81 & 82.50 & 42.30 & 62.60 \\
\midrule
\multirow{8}{*}{\makecell[c]{Qwen3\\8B\\Base}}
& Base & 74.98 & 90.00 & \textbf{31.85} & 53.68 & 39.94 & 59.56 & 12.92 & 33.33 & 53.75 & 80.00 & 42.69 & 63.31 \\
& SFT & 70.28 & 87.60 & 21.92 & 34.93 & 33.07 & 56.44 & 10.83 & 20.00 & 45.94 & 82.50 & 36.41 & 56.29 \\
& DFT & 74.92 & 87.00 & 22.93 & 30.51 & 39.61 & 57.33 & 16.67 & 30.00 & 56.56 & 87.50 & 42.14 & 58.47 \\
& ASFT & 76.55 & 89.80 & 23.48 & 31.62 & 41.35 & 60.30 & 16.25 & 26.67 & 61.56 & 85.00 & 43.84 & 58.68 \\
& TALR & 72.42 & 88.00 & 23.12 & 36.76 & 36.78 & 56.74 & 14.58 & 26.67 & 49.06 & 85.00 & 39.19 & 58.63 \\
& PriFT & 80.47 & 92.00 & 30.56 & 39.71 & 45.89 & 64.74 & \textbf{23.75} & 43.33 & 62.50 & 90.00 & 48.63 & 65.96 \\
& SPFT & 80.00 & 92.20 & 28.08 & 38.97 & 45.07 & 63.26 & 20.83 & 40.00 & 65.00 & 90.00 & 47.80 & 64.89 \\
& \textbf{SPFT-SR} & \textbf{81.10} & \textbf{92.60} & 29.64 & 40.81 & \textbf{46.50} & \textbf{65.93} & 23.33 & \textbf{46.67} & \textbf{68.12} & \textbf{92.50} & \textbf{49.74} & \textbf{67.70} \\
\bottomrule
\end{tabular}}
\end{table*}
\paragraph{Mathematical Results.}
Table~\ref{tab:math_main} summarizes the main mathematical results. Each benchmark reports Avg@8 and Pass@8, and the final pair reports the five-benchmark average Avg@8 and Pass@8.
SPFT-SR improves over SFT on both metrics at the two main reported model scales.

\begin{figure*}[t]
\centering
\begin{subfigure}[t]{0.48\textwidth}
  \includegraphics[width=\linewidth]{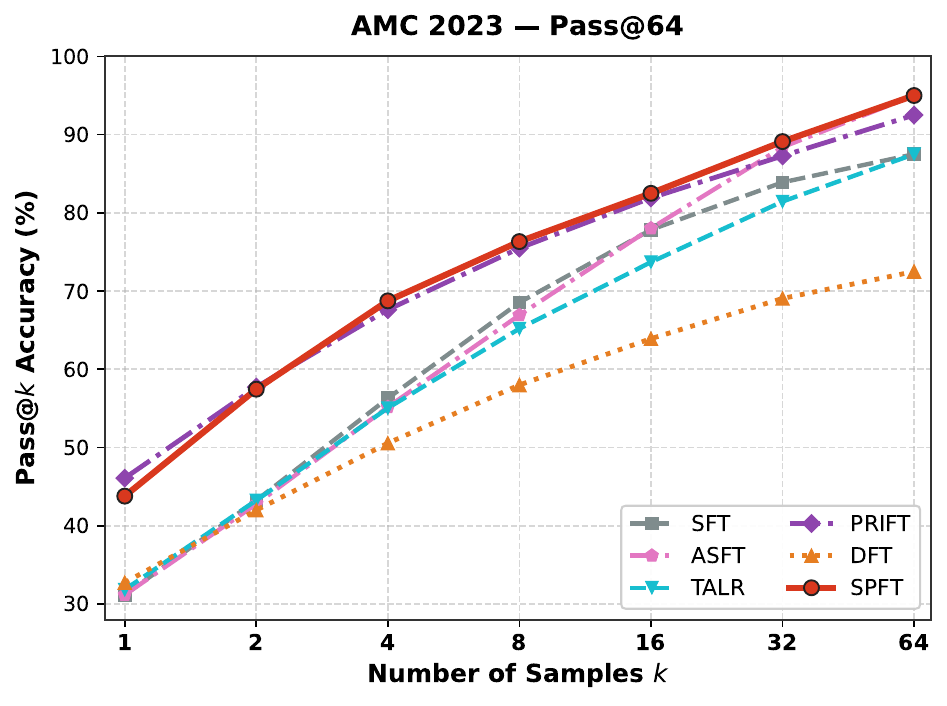}
  \caption{Pass@k}
\end{subfigure}\hfill
\begin{subfigure}[t]{0.48\textwidth}
  \includegraphics[width=\linewidth]{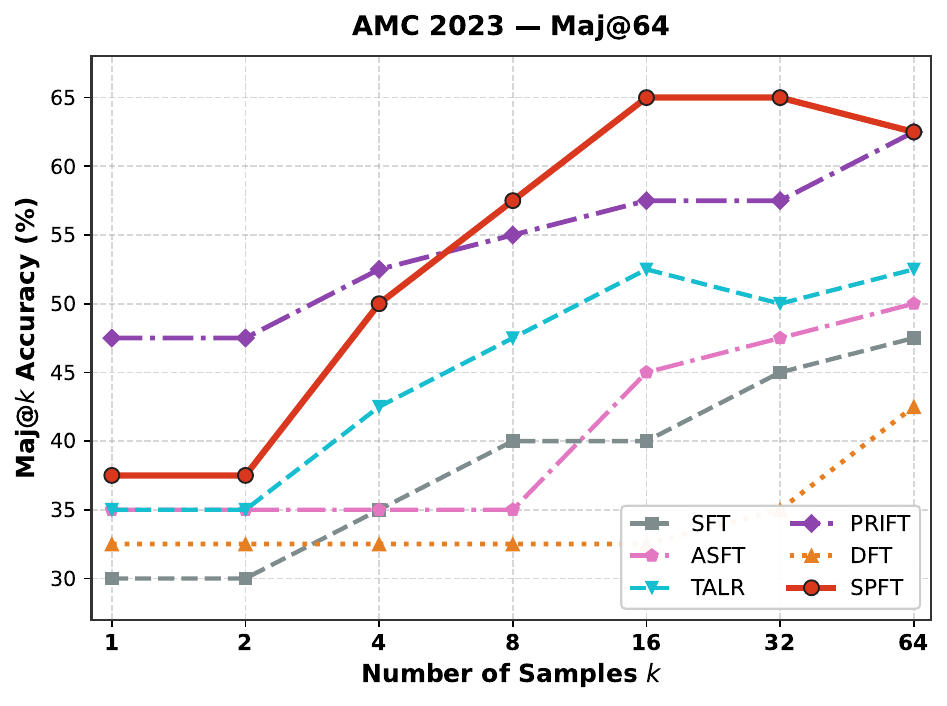}
  \caption{Maj@k}
\end{subfigure}
\caption{Response-level sampling scaling on AMC 2023. Increasing $k$ reveals whether a method can obtain a correct solution from multiple samples (Pass@k) and whether correct solutions become sufficiently dominant for majority voting (Maj@k). SPFT combines high discovery probability with strong consensus across the sampling budget.}
\label{fig:amc23_sampling_scaling}
\end{figure*}

The response-level sampling curves in Figure~\ref{fig:amc23_sampling_scaling} complement the token-level diagnostics in Figure~\ref{fig:weighting_dynamics_metrics}. Pass@k measures whether at least one of $k$ sampled solutions is correct, while Maj@k measures correctness under majority voting. SPFT maintains strong growth in both metrics as $k$ increases, indicating that it preserves useful output diversity while also increasing agreement around correct solutions. This is evidence for controlled sharpening rather than the near-deterministic collapse induced by DFT; Pass@k and Maj@k are response-level diagnostics, not direct token-level entropy measurements.

\label{sec:exp_code}

\begin{table}[t]
\centering
\caption{Qwen2.5-Coder-1.5B EvalPlus results. All values are greedy Pass@1 percentages; $\Delta$ reports the change in four-task Code Avg relative to SFT (49.03\%).}
\label{tab:code_main}
\resizebox{\linewidth}{!}{
\begin{tabular}{lrrrrrr}
\toprule
\textbf{Method} & \textbf{HumanEval} & \textbf{HumanEval+} & \textbf{MBPP} & \textbf{MBPP+} & \textbf{Code Avg} & \textbf{$\Delta$ vs SFT} \\
\midrule
Base & 43.9 & 36.6 & 69.2 & 58.6 & 52.08 & -- \\
SFT & 41.5 & 37.2 & 63.2 & 54.2 & 49.03 & 0.00 \\
ASFT & 49.4 & 43.3 & 67.2 & 55.8 & 53.92 & +4.89 \\
PriFT & 56.1 & 50.6 & 67.7 & 57.4 & 57.95 & +8.92 \\
TALR & 59.8 & 52.4 & 67.5 & 56.3 & 59.00 & +9.97 \\
\textbf{SPFT} & \textbf{56.7} & \textbf{53.0} & \textbf{70.6} & \textbf{58.5} & \textbf{59.70} & \textbf{+10.67} \\
\bottomrule
\end{tabular}
}
\end{table}

\paragraph{Code Results.} Table~\ref{tab:code_main} reports the Qwen2.5-Coder-1.5B comparison. The final column reports the improvement in four-task Code Avg relative to SFT. SPFT achieves the best four-task Code Avg among the trained methods, 59.70\%, improving over SFT by 10.67 percentage points.

\section{Ablation Studies}
\label{sec:ablations}

We use two ablations in this section to isolate the design choices behind SPFT. We first test whether the privileged reference must remain frozen and then vary the form of privileged information in code generation. A complementary comparison between scalar target-token weighting, as in DFT~\citep{wu2025generalization}, and full-vocabulary knowledge distillation~\citep{hinton2015distilling} is reported in Appendix~\ref{app:kd_ablation}.

\subsection{Frozen versus Online Privileged Reference}
\label{sec:ablation_frozen_online}

Our theoretical analysis identifies the reference weight as part of the optimization target rather than as a detached implementation detail. With a frozen privileged reference model, the weight is fixed throughout training:
\begin{equation}
    q_t^{\mathrm{frozen}}
    =
    p_{\theta_0}(y_t \mid x,c,y_{<t})
    \label{eq:ablation_frozen_weight}
\end{equation}
The corresponding static objective preserves a stable reference prior, so the model is optimized against the same privileged signal at every update. The online variant removes this invariance by evaluating the privileged branch with the current parameters:
\begin{equation}
    q_t^{\mathrm{online}}
    =
    \operatorname{sg}\! \left[ p_{\theta}(y_t \mid x,c,y_{<t}) \right].
    \label{eq:ablation_online_weight}
\end{equation}
The reference weight then updates as the model optimizes. As established in Section~\ref{sec:theory}, a frozen reference provides a stationary statistical target that ensures stable optimization. Table \ref{tab:ablation_frozen_online} shows the comparison between the two variants. The consistent aggregate improvement supports the theoretical prediction that a fixed reference prior is more stable than a reference that evolves dynamically alongside the model during optimization.

\begin{table*}[t]
\centering
\small
\setlength{\tabcolsep}{4.2pt}
\renewcommand{\arraystretch}{1.08}
\caption{Frozen versus online privileged reference for Qwen2.5-Math-1.5B. Each benchmark reports Avg@8 / Pass@8, and the final pair gives the unweighted five-benchmark average.}
\label{tab:ablation_frozen_online}
\vspace{0.3em}
\resizebox{\textwidth}{!}{\begin{tabular}{clcccccccccccc}
\toprule
\multirow{2}{*}{\textbf{Reference}} &
& \multicolumn{2}{c}{\textbf{MATH-500}}
& \multicolumn{2}{c}{\textbf{Minerva Math}}
& \multicolumn{2}{c}{\textbf{OlympiadBench}}
& \multicolumn{2}{c}{\textbf{AIME24}}
& \multicolumn{2}{c}{\textbf{AMC23}}
& \multicolumn{2}{c}{\textbf{Average}} \\
\cmidrule(lr){3-4}\cmidrule(lr){5-6}\cmidrule(lr){7-8}\cmidrule(lr){9-10}\cmidrule(lr){11-12}\cmidrule(lr){13-14}
& & Avg@8 & Pass@8 & Avg@8 & Pass@8 & Avg@8 & Pass@8 & Avg@8 & Pass@8 & Avg@8 & Pass@8 & Avg@8 & Pass@8 \\
\midrule
Online ($\theta_t$) & & 60.30 & 84.00 & 21.65 & 51.10 & 23.61 & 47.70 & 5.42 & 16.67 & 38.12 & 75.00 & 29.82 & 54.89 \\
\textbf{Frozen ($\theta_0$)} & & \textbf{64.75} & \textbf{87.40} & 20.82 & 40.44 & \textbf{28.94} & \textbf{51.26} & \textbf{7.08} & 16.67 & \textbf{43.12} & \textbf{82.50} & \textbf{32.94} & \textbf{55.65} \\
\midrule
Frozen $-$ Online & & +4.45 & +3.40 & -0.83 & -10.66 & +5.33 & +3.56 & +1.66 & 0.00 & +5.00 & +7.50 & \textbf{+3.12} & \textbf{+0.76} \\
\bottomrule
\end{tabular}}
\end{table*}

\subsection{Privileged-Information Form Ablation}
\label{sec:ablation_pi_form}

The main SPFT code result uses Clean Rule as privileged information, which conveys algorithmic intent without exposing the reference implementation. More details are provided in Section~\ref{app:code_clean_rule}. To isolate the role of the privileged-information representation, we compare Clean Rule against three alternatives: the complete reference code, unit-test information, and a suffix-half fragment of the reference code. Full-code conditioning supplies the most direct answer-level signal but can leak surface syntax and boilerplate. Unit-test information provides outcome-related constraints without exposing the entire implementation. Suffix-Half provides a partial textual reference and tests whether a local code fragment is sufficient. Clean Rule instead supplies a compact abstraction of algorithmic paradigms, data structures, and boundary guards. Table~\ref{tab:code_pi_ablation} shows that Clean Rule achieves the highest Code Avg score, which supports the use of high-level algorithmic guidance rather than direct code leakage.

\begin{table}[t]
\centering
\caption{Privileged-information form ablation for Qwen2.5-Coder-1.5B. All rows use the same SPFT training objective; only the privileged prompt content changes.}
\label{tab:code_pi_ablation}
\resizebox{\linewidth}{!}{
\begin{tabular}{lrrrrr}
\toprule
\textbf{Privileged information} & \textbf{HumanEval} & \textbf{HumanEval+} & \textbf{MBPP} & \textbf{MBPP+} & \textbf{Code Avg} \\
\midrule
SPFT (Clean Rule) & 56.7 & 53.0 & 70.6 & 58.5 & 59.70 \\
SPFT (Full Code) & 58.5 & 54.3 & 63.9 & 54.1 & 57.70 \\
SPFT (Unit Tests) & 53.0 & 48.2 & 68.2 & 56.6 & 56.50 \\
SPFT (Suffix-Half) & 57.3 & 51.8 & 66.2 & 56.9 & 58.05 \\
\bottomrule
\end{tabular}
}
\end{table}

\section{Conclusion and Limitations}
\label{sec:conclusion}

We presented Self-Privileged Fine-Tuning (SPFT), an offline token-reweighted SFT framework that uses privileged information available during training but hidden from the model at inference. A frozen copy of the initial model conditions on privileged information, such as a verified answer or a structured description of algorithmic intent, and assigns scalar weights to target tokens while the trainable model receives only the ordinary task prompt. This design provides outcome-aware supervision without an external teacher model or online rollouts. We further developed a theoretical framework that distinguishes dynamic and static weighting, showing why a frozen reference is needed to preserve a fixed optimization target and avoid the moving, self-reinforcing signal caused by online updates. Experiments on mathematical reasoning and code generation, together with targeted ablations, demonstrate the effectiveness of SPFT and support these design choices.

SPFT currently relies on useful verifiable or structurally extractable privileged information, and our evaluation is limited to mathematics and code. Future work should investigate how to construct and validate appropriate privileged information in other domains, especially when outcomes are noisy, partial, subjective, or long-horizon. Extending SPFT to broader tasks and modalities, while improving robustness to imperfect privileged signals, is an important direction.

\section*{Ethics Statement}
This work does not raise significant ethical concerns beyond those commonly associated with research on language-model training. It does not involve human subjects, private or sensitive personal data, or deployment in high-stakes settings. We use publicly available datasets and open-source models, and we did not identify additional ethical risks specific to the methods or experiments presented here. The authors take responsibility for the ethical conduct and content of this work.

\section*{AI Assistance Disclosure}
With the exception of limited implementation-related code assistance described below, we did not use generative AI tools to generate synthetic datasets, develop theoretical models or conceptual frameworks, formulate mathematical claims, provide critical ingredients for proofs, assist in writing proofs, propose or refine hypotheses, design or provide feedback on the research methodology or experiments, translate content, clean or reformat datasets, perform qualitative or thematic analysis, or analyze and interpret experimental results. Generative AI tools were used to assist with manuscript drafting and editing, including improving readability, grammar, spelling, and phrasing. They were also used in a limited capacity during implementation, primarily to identify potential bugs and vulnerabilities; the method design and algorithmic decisions were made by the authors. The authors reviewed all AI-assisted text for accuracy and consistency and verified the resulting implementation through manual inspection and testing. The authors take full responsibility for the final content of this paper, including all AI-assisted contributions.

\section*{Reproducibility Statement}
The assumptions and theoretical results are described in Section~3 and the appendix. The SPFT method and its implementation details are presented in Section~4, while the datasets, hyperparameters, training configurations, baselines, and evaluation protocols are described in Sections~5--6 and the appendix. These details are provided to facilitate reproduction of our results.

\section*{Acknowledgments}
The authors acknowledge support from The Hong Kong Polytechnic University (Research Grant P0046811).

\bibliography{iclr2027_conference}
\bibliographystyle{iclr2027_conference}

\newpage
\appendix

\section{Mathematical Proofs and Weighting Implementation}
\label{app:proofs}

\subsection{Proof of Proposition~\ref{prop:dft_limit} (Stepwise Optimum and Limiting Distribution of DFT)}
\label{app:proof_prop1}

For a fixed prefix context $(x, y_{<t})$, let $p_{\mathrm{data}}(v \mid x, y_{<t})$ denote the conditional next-token distribution of the demonstration dataset over the vocabulary $\mathcal{V}$, and let $p_{\theta_k}(v \mid x, y_{<t})$ denote the model's distribution at optimization step $k$. At step $k$, the DFT objective in Eq.~\ref{eq:dft_step_objective} decomposes across contexts $(x, y_{<t})$. For a given context, minimizing the expected weighted negative log-likelihood with respect to the next-token distribution $p_{\theta}(\cdot \mid x, y_{<t})$ corresponds to solving:
\begin{equation}
    \min_{p_{\theta}(\cdot \mid x, y_{<t}) \in \Delta(\mathcal{V})}
    -\sum_{v \in \mathcal{V}}
    p_{\mathrm{data}}(v \mid x, y_{<t}) \, p_{\theta_k}(v \mid x, y_{<t})
    \log p_{\theta}(v \mid x, y_{<t}).
\end{equation}
Define the unnormalized nonnegative weight $u_k(v) \triangleq p_{\mathrm{data}}(v \mid x, y_{<t}) \, p_{\theta_k}(v \mid x, y_{<t})$ and the partition constant $Z_k(x, y_{<t}) = \sum_{v' \in \mathcal{V}} u_k(v')$. Provided $Z_k(x, y_{<t}) > 0$, normalizing $u_k$ yields a valid probability distribution $\tilde{p}_k(v) = u_k(v) / Z_k(x, y_{<t})$. By Gibbs' inequality (nonnegativity of KL divergence),
\begin{equation}
    -\sum_{v \in \mathcal{V}} u_k(v) \log p_{\theta}(v \mid x, y_{<t})
    = Z_k(x, y_{<t}) \Bigl[
        D_{\mathrm{KL}}\bigl(\tilde{p}_k(\cdot) \parallel p_{\theta}(\cdot \mid x, y_{<t})\bigr)
        + H(\tilde{p}_k)
    \Bigr],
\end{equation}
which is uniquely minimized when $p_{\theta_{k+1}}(v \mid x, y_{<t}) = \tilde{p}_k(v) \propto p_{\mathrm{data}}(v \mid x, y_{<t}) \, p_{\theta_k}(v \mid x, y_{<t})$, establishing Eq.~\ref{eq:dft_step_optimum}.

By induction over $k \ge 0$, repeatedly applying Eq.~\ref{eq:dft_step_optimum} yields the closed-form iterate in Eq.~\ref{eq:dft_closed_form}:
\begin{equation}
    p_{\theta_k}(v \mid x, y_{<t})
    = \frac{p_{\theta_0}(v \mid x, y_{<t}) \, p_{\mathrm{data}}(v \mid x, y_{<t})^k}
    {\sum_{v' \in \mathcal{V}} p_{\theta_0}(v' \mid x, y_{<t}) \, p_{\mathrm{data}}(v' \mid x, y_{<t})^k}.
\end{equation}
Let $v^* = \arg\max_{v \in \mathcal{V}} p_{\mathrm{data}}(v \mid x, y_{<t})$ be the unique mode of $p_{\mathrm{data}}(\cdot \mid x, y_{<t})$, with $p_{\theta_0}(v^* \mid x, y_{<t}) > 0$. Dividing both numerator and denominator by $p_{\mathrm{data}}(v^* \mid x, y_{<t})^k$ gives:
\begin{equation}
    p_{\theta_k}(v \mid x, y_{<t})
    = \frac{p_{\theta_0}(v \mid x, y_{<t}) \left(\frac{p_{\mathrm{data}}(v \mid x, y_{<t})}{p_{\mathrm{data}}(v^* \mid x, y_{<t})}\right)^k}
    {p_{\theta_0}(v^* \mid x, y_{<t}) + \sum_{v' \neq v^*} p_{\theta_0}(v' \mid x, y_{<t}) \left(\frac{p_{\mathrm{data}}(v' \mid x, y_{<t})}{p_{\mathrm{data}}(v^* \mid x, y_{<t})}\right)^k}.
\end{equation}
Since $v^*$ is the unique maximizer, every $v' \neq v^*$ satisfies $p_{\mathrm{data}}(v' \mid x, y_{<t}) / p_{\mathrm{data}}(v^* \mid x, y_{<t}) < 1$, so the corresponding ratio raised to the $k$-th power converges to $0$ as $k \to \infty$. Consequently, $\lim_{k \to \infty} p_{\theta_k}(v \mid x, y_{<t}) = \mathbf{1}[v = v^*]$, completing the proof of Proposition~\ref{prop:dft_limit}. $\hfill\blacksquare$

\subsection{Proof of Proposition~\ref{prop:static_optimum} (Optimal Distribution of Static Weighting)}
\label{app:proof_prop2}

Consider the static token-weighted fine-tuning objective in Eq.~\ref{eq:static_objective}, where $p_{\mathrm{ref}}(v \mid x, y_{<t})$ is a frozen reference distribution independent of $\theta$:
\begin{equation}
    \mathcal{L}_{\mathrm{static}}(\theta)
    = -\mathbb{E}_{(x, y) \sim p_{\mathrm{data}}}
    \left[
        \sum_t p_{\mathrm{ref}}(y_t \mid x, y_{<t}) \log p_{\theta}(y_t \mid x, y_{<t})
    \right].
\end{equation}
Conditioned on any fixed prefix $(x, y_{<t})$ in the support of $p_{\mathrm{data}}$, the pointwise expected loss over the next token $v \in \mathcal{V}$ is:
\begin{equation}
    \ell_{\mathrm{static}}\bigl(p_\theta(\cdot \mid x, y_{<t})\bigr)
    = -\sum_{v \in \mathcal{V}} p_{\mathrm{data}}(v \mid x, y_{<t}) \, p_{\mathrm{ref}}(v \mid x, y_{<t}) \log p_\theta(v \mid x, y_{<t}).
\end{equation}
Let $Z_{\mathrm{ref}}(x, y_{<t}) = \sum_{v' \in \mathcal{V}} p_{\mathrm{data}}(v' \mid x, y_{<t}) \, p_{\mathrm{ref}}(v' \mid x, y_{<t})$ denote the overlap partition constant. Defining the normalized product distribution
\begin{equation}
    p_\theta^*(v \mid x, y_{<t})
    = \frac{p_{\mathrm{data}}(v \mid x, y_{<t}) \, p_{\mathrm{ref}}(v \mid x, y_{<t})}{Z_{\mathrm{ref}}(x, y_{<t})},
\end{equation}
we can rewrite the pointwise objective up to an additive constant independent of $\theta$ as:
\begin{align}
    \ell_{\mathrm{static}}\bigl(p_\theta(\cdot \mid x, y_{<t})\bigr)
    &= Z_{\mathrm{ref}}(x, y_{<t}) \, D_{\mathrm{KL}}\!\left(p_\theta^*(\cdot \mid x, y_{<t}) \;\middle\|\; p_\theta(\cdot \mid x, y_{<t})\right) \nonumber \\
    &\quad + Z_{\mathrm{ref}}(x, y_{<t}) \, H\bigl(p_\theta^*(\cdot \mid x, y_{<t})\bigr).
\end{align}
By Gibbs' inequality, the KL divergence is uniquely minimized at zero when $p_\theta(v \mid x, y_{<t}) = p_\theta^*(v \mid x, y_{<t}) \propto p_{\mathrm{data}}(v \mid x, y_{<t}) \, p_{\mathrm{ref}}(v \mid x, y_{<t})$. Because $p_{\mathrm{ref}}$ is held fixed across optimization steps, $p_\theta^*$ is stationary throughout training. $\hfill\blacksquare$

\subsection{Proof of Proposition~\ref{prop:spft_ratio} and Connection to Outcome-Conditioned Inference}
\label{app:proof_prop3}

Let $(x, y, c)$ denote a problem prompt $x$, a demonstration sequence $y = (y_1, \dots, y_T)$, and the verified outcome $c$. Under the frozen reference model $p_{\theta_0}$, applying Bayes' rule to the sequence-level conditional probability $p_{\theta_0}(y \mid x, c)$ gives:
\begin{equation}
    p_{\theta_0}(y \mid x, c)
    = \frac{p_{\theta_0}(y, c \mid x)}{p_{\theta_0}(c \mid x)}
    = \frac{p_{\theta_0}(c \mid x, y)}{p_{\theta_0}(c \mid x)} \, p_{\theta_0}(y \mid x),
\end{equation}
which establishes the sequence-level ratio relation. Similarly, conditioning on the prefix $y_{<t}$ at step $t \in \{1, \dots, T\}$ and applying Bayes' rule to the event $(y_t, c)$ given $(x, y_{<t})$ yields:
\begin{align}
    q_t \;\triangleq\; p_{\theta_0}(y_t \mid x, c, y_{<t})
    &= \frac{p_{\theta_0}(y_t, c \mid x, y_{<t})}{p_{\theta_0}(c \mid x, y_{<t})} \nonumber \\
    &= \frac{p_{\theta_0}(c \mid x, y_{\le t})}{p_{\theta_0}(c \mid x, y_{<t})} \, p_{\theta_0}(y_t \mid x, y_{<t})
    = \frac{p_{\theta_0}(c \mid x, y_{\le t})}{p_{\theta_0}(c \mid x, y_{<t})} \, p_t^{\mathrm{prior}},
\end{align}
which establishes Eq.~\ref{eq:spft_token_ratio}.

\subsection{Implementation and Interpretation of SPFT-SR Weights}
\label{app:sr_implementation}
\label{app:sr_interpretation}

As introduced in Section~\ref{sec:method_sr}, SPFT-SR operates on detached privileged reference probabilities $q_t = p_{\theta_0}(y_t \mid x, c, y_{<t})$ rather than log-probabilities. Consider one sample $(x,y,c)$ with $T$ valid response tokens. To compare privileged confidence within this response, we compute
\begin{equation}
    \mu
    =
    \frac{1}{T}\sum_{t=1}^{T}q_t,
    \qquad
    \sigma
    =
    \sqrt{
    \frac{1}{T}\sum_{t=1}^{T}(q_t-\mu)^2
    + \epsilon
    },
    \label{eq:app_sr_stats}
\end{equation}
where $\epsilon = 10^{-8}$ ensures numerical stability. The clipped sequence-relative score is
\begin{equation}
    r_t
    =
    \left[\frac{q_t-\mu}{\sigma}\right]_+.
    \label{eq:app_sr_clipped}
\end{equation}
Only tokens with above-average privileged confidence receive positive scores. We rescale them to have mean weight one over the response:
\begin{equation}
    w_t
    =
    \frac{r_t}
    {\max\!\left(\frac{1}{T}\sum_{u=1}^{T}r_u,\,\epsilon\right)}.
    \label{eq:app_sr_rescaled}
\end{equation}
When $\sum_{u=1}^{T}r_u\leq\epsilon$, the implementation instead uses uniform weights $w_t=1$ to avoid a zero-gradient update. The resulting weights are detached before multiplying the model's token-level cross-entropy loss. This is a sequence-relative calibration transformation, not an exact estimator of a log-ratio soft advantage.

\section{Privileged Information Construction and Conditioning Interfaces}
\label{app:prompts}

This section provides the exact prompt templates, static AST extraction rules, and concrete walkthrough examples for both mathematical reasoning and program synthesis, as well as the exact specifications of all four privileged-information (PI) variants evaluated in the code ablation study (Section~\ref{sec:ablation_pi_form}, Table~\ref{tab:code_pi_ablation}).

\subsection{Mathematical Reasoning Conditioning Interface (\texttt{opsd\_answer})}
\label{app:math_pi_template}

In mathematical reasoning on NuminaMath-CoT 10K, the ground-truth terminal answer $c$ is extracted from the final \texttt{\textbackslash boxed\{...\}} expression of the verified solution $y$. The frozen privileged reference model receives $c$ via the \texttt{opsd\_answer} template, whereas the target model receives only the standard problem prompt:
\begin{verbatim}
Problem: {problem}

Here is a correct reference final answer to this problem:
=== Reference Answer Begin ===
{ground_truth_answer}
=== Reference Answer End ===

Use the reference final answer only as privileged guidance.
Independently derive a valid solution; do not merely copy
the answer. Think step by step, explore different approaches,
and don't be afraid to backtrack or reconsider if something
doesn't work out:

Please reason step by step, and put your final answer
within \boxed{}.
\end{verbatim}

\subsection{Program Synthesis Privileged Information: Clean Rule Specification}
\label{app:code_clean_rule}

\paragraph{Motivation.}
A complete reference implementation is an overly detailed privileged signal: it exposes function signatures, variable names, formatting, and routine control-flow tokens in addition to the algorithmic idea. Such surface-form information can make the reference model highly confident on boilerplate tokens and dilute the relative emphasis placed on algorithmic decisions. Clean Rule PI instead provides a compact description of the solution's structure without supplying the implementation itself.

\paragraph{High-level construction.}
Clean Rule PI is obtained by lightweight deterministic static analysis of the verified reference solution. The analyzer identifies broad algorithmic paradigms (for example, dynamic programming, search, graph traversal, greedy or heap-based processing, hashing, sorting, and bit manipulation), records the main abstract data structures, and extracts a small number of boundary or early-return conditions when present. These signals are deduplicated and formatted as a short structural descriptor with three possible fields: \emph{Paradigms}, \emph{Key Data Structures}, and \emph{Boundary Invariants}. The extraction is offline, deterministic, and does not call an external language model. The resulting descriptor is supplied only to the frozen privileged reference model; the target model and the inference-time model receive the original problem without the privileged description.

\paragraph{Concrete example.}
The following example illustrates the transformation from a verified implementation to its abstract privileged signal. The implementation is used only as the source for extraction; the privileged reference model receives only the second item.
\begin{itemize}
    \item \textbf{Ground-truth reference:}
\begin{verbatim}
def twoSum(nums, target):
    seen = {}
    for i, num in enumerate(nums):
        comp = target - num
        if comp in seen:
            return [seen[comp], i]
        seen[num] = i
    return []
\end{verbatim}
    \item \textbf{Extracted Clean Rule PI:}
\begin{verbatim}
Paradigms: Iterative Simulation
Key Data Structures: Hash Map / Set
Boundary Invariants: Checked for ['comp in seen']
\end{verbatim}
\end{itemize}

\section{Additional Ablation Studies}
\label{app:additional_ablations}

\subsection{Direct Token Weighting versus Full-Vocabulary KD}
\label{app:kd_ablation}

SPFT uses the privileged reference model only to assign a scalar weight to the verified target token, as in token-reweighted objectives such as DFT~\citep{wu2025generalization}. Standard knowledge distillation instead matches the reference and target distributions over the entire vocabulary at every prefix~\citep{hinton2015distilling}. For a valid response position $t$, let
\begin{equation}
    q_t(v)=p_{\theta_0}(v\mid x,c,y_{<t}),
    \qquad
    p_t(v)=p_\theta(v\mid x,y_{<t}).
    \label{eq:ablation_vocab_distributions}
\end{equation}
A full-vocabulary forward-KL control minimizes
\begin{equation}
    \mathcal{L}_{\mathrm{KD}}^{\rightarrow}(\theta;x,y,c)
    =
    \sum_{t=1}^{T}
    D_{\mathrm{KL}}\!\left(q_t(\cdot)\,\middle\|\,p_t(\cdot)\right),
    \label{eq:ablation_forward_kl}
\end{equation}
while the reverse-KL direction provides a complementary divergence control:
\begin{equation}
    \mathcal{L}_{\mathrm{KD}}^{\leftarrow}(\theta;x,y,c)
    =
    \sum_{t=1}^{T}
    D_{\mathrm{KL}}\!\left(p_t(\cdot)\,\middle\|\,q_t(\cdot)\right).
    \label{eq:ablation_reverse_kl}
\end{equation}
Both controls require matching a complete vocabulary distribution at every training prefix. This creates a potential train--test mismatch: training observes prefixes from the verified response through teacher forcing, whereas inference generates prefixes from the model's own policy. Once a generated prefix deviates from the reference trajectory, the model is evaluated in contexts that were not represented by the KD loss.
SPFT has a narrower supervision interface. It evaluates the privileged reference model on the same offline reference prefix but transfers only a scalar importance signal to the observed target token. As established in Section~\ref{sec:theory}, this weighted SFT objective induces a target distribution proportional to the product of the data distribution and the reference distribution. It therefore avoids requiring the model to reproduce the reference's entire conditional distribution on each prefix.

\begin{table*}[t]
\centering
\small
\setlength{\tabcolsep}{4.5pt}
\renewcommand{\arraystretch}{1.12}
\caption{Direct token weighting versus full-vocabulary KL for Qwen2.5-Math-1.5B. Each benchmark reports Avg@8 / Pass@8, and the final pair gives the unweighted five-benchmark average.}
\label{tab:ablation_full_kl}
\resizebox{\textwidth}{!}{\begin{tabular}{clcccccccccccc}
\toprule
\multirow{2}{*}{\textbf{Objective}} &
& \multicolumn{2}{c}{\textbf{MATH-500}}
& \multicolumn{2}{c}{\textbf{Minerva Math}}
& \multicolumn{2}{c}{\textbf{OlympiadBench}}
& \multicolumn{2}{c}{\textbf{AIME24}}
& \multicolumn{2}{c}{\textbf{AMC23}}
& \multicolumn{2}{c}{\textbf{Average}} \\
\cmidrule(lr){3-4}\cmidrule(lr){5-6}\cmidrule(lr){7-8}\cmidrule(lr){9-10}\cmidrule(lr){11-12}\cmidrule(lr){13-14}
& & Avg@8 & Pass@8 & Avg@8 & Pass@8 & Avg@8 & Pass@8 & Avg@8 & Pass@8 & Avg@8 & Pass@8 & Avg@8 & Pass@8 \\
\midrule
SPFT & & \textbf{64.75} & \textbf{87.40} & \textbf{20.82} & \textbf{40.44} & \textbf{28.94} & \textbf{51.26} & \textbf{7.08} & \textbf{16.67} & \textbf{43.12} & \textbf{82.50} & \textbf{32.94} & \textbf{55.65} \\
Forward KL & & 41.23 & 81.80 & 9.97 & 37.50 & 19.02 & 46.37 & 3.75 & 16.67 & 20.00 & 65.00 & 18.79 & 49.47 \\
Reverse KL & & 31.55 & 76.40 & 8.36 & 32.35 & 15.96 & 44.59 & 1.67 & 6.67 & 16.88 & 60.00 & 14.88 & 44.00 \\
\bottomrule
\end{tabular}}
\end{table*}

Table~\ref{tab:ablation_full_kl} shows a large gap between direct token weighting and full-vocabulary distribution matching. Forward KL achieves average Avg@8 / Pass@8 scores of 18.79\% / 49.47\%, while reverse KL achieves 14.88\% / 44.00\%, compared with 32.94\% / 55.65\% for SPFT. Thus, the complete-vocabulary objectives lose 14.15 / 6.18 and 18.06 / 11.65 percentage points in Avg@8 / Pass@8, respectively.

\section{Complete Experimental Details and Extended Results}
\label{app:math_majority}

\subsection{Training and Evaluation Hyperparameters}
\label{app:hyperparameters}

Table~\ref{tab:hyperparams} summarizes the training and evaluation configurations used across all mathematical reasoning and code generation experiments in Section~\ref{sec:experiments} and Section~\ref{sec:ablations}. All methods within each domain share identical training data, batch sizes, learning rate schedules, sequence lengths, and evaluation decoding protocols.

\begin{table}[ht]
\centering
\caption{Summary of training and evaluation hyperparameters for mathematical reasoning and code generation.}
\label{tab:hyperparams}
\small
\resizebox{\linewidth}{!}{\begin{tabular}{lll}
\toprule
\textbf{Configuration Item} & \textbf{Mathematical Reasoning} & \textbf{Code Generation} \\
\midrule
Training dataset & NuminaMath-CoT 10K (9,637 verified pairs) & OpenCodeInstruct 10K (10,000 verified pairs) \\
Backbone models & Qwen2.5-Math-1.5B, Qwen2.5-Math-7B, Qwen3-8B-Base & Qwen2.5-Coder-1.5B \\
Privileged prompt style ($c$) & \texttt{opsd\_answer} (boxed final answer) & \texttt{code\_rule\_paradigm} (Clean Rule AST summary) \\
Global batch size & 32 & 16 \\
Peak learning rate & $1.0 \times 10^{-5}$ & $5.0 \times 10^{-5}$ \\
LR schedule \& warmup & Cosine decay, 5\% linear warmup & Cosine decay, 5\% linear warmup \\
Training duration & 1 epoch (301 optimization steps) & 1 epoch (625 optimization steps) \\
Maximum sequence length & 3,072 tokens & 2,048 tokens \\
Optimizer \& precision & AdamW ($\beta_1=0.9, \beta_2=0.999$), BF16 + FSDP & AdamW ($\beta_1=0.9, \beta_2=0.999$), BF16 + FSDP \\
Evaluation benchmarks & MATH-500, Minerva, OlympiadBench, AIME24, AMC23 & EvalPlus: HumanEval, HumanEval+, MBPP, MBPP+ \\
Evaluation decoding & $n=8$ samples, temperature $0.7$, top-$p$ $0.8$, seed 0 & Greedy decoding ($T=0$, Pass@1) \\
Reported metrics & Avg@8, Pass@8 (Table~\ref{tab:math_main}); Maj@8 (Table~\ref{tab:math_majority}) & Pass@1 on 4 tasks and unweighted Code Avg (Table~\ref{tab:code_main}) \\
\bottomrule
\end{tabular}}
\end{table}

\subsection{Additional Model-Scale Results}
\label{app:additional_math_scales}

Table~\ref{tab:additional_math_scales} uses the same per-benchmark format as Table~\ref{tab:math_main}. Each benchmark reports Avg@8 / Pass@8, and the final pair gives the unweighted five-benchmark average.

\begin{table*}[t]
\centering
\small
\setlength{\tabcolsep}{4.5pt}
\renewcommand{\arraystretch}{1.12}
\caption{Additional mathematical reasoning results in the same format as Table~\ref{tab:math_main}.}
\label{tab:additional_math_scales}
\resizebox{\textwidth}{!}{\begin{tabular}{clcccccccccccc}
\toprule
\multirow{2}{*}{\textbf{Model}} & \multirow{2}{*}{\textbf{Method}}
& \multicolumn{2}{c}{\textbf{MATH-500}}
& \multicolumn{2}{c}{\textbf{Minerva Math}}
& \multicolumn{2}{c}{\textbf{OlympiadBench}}
& \multicolumn{2}{c}{\textbf{AIME24}}
& \multicolumn{2}{c}{\textbf{AMC23}}
& \multicolumn{2}{c}{\textbf{Average}} \\
\cmidrule(lr){3-4}\cmidrule(lr){5-6}\cmidrule(lr){7-8}\cmidrule(lr){9-10}\cmidrule(lr){11-12}\cmidrule(lr){13-14}
& & Avg@8 & Pass@8 & Avg@8 & Pass@8 & Avg@8 & Pass@8 & Avg@8 & Pass@8 & Avg@8 & Pass@8 & Avg@8 & Pass@8 \\
\midrule
\multirow{8}{*}{\makecell[c]{Qwen2.5-Math\\1.5B}}
& Base & 44.15 & 82.20 & 8.04 & 26.84 & 23.81 & 50.37 & 6.25 & 13.33 & 25.00 & 67.50 & 21.45 & 48.05 \\
& SFT & 55.88 & 82.40 & 15.40 & 37.87 & 22.37 & 49.48 & 5.00 & 23.33 & 31.25 & 60.00 & 25.98 & 50.62 \\
& DFT & 62.92 & 80.00 & 22.24 & 38.60 & 24.41 & 43.56 & 4.58 & 10.00 & 34.06 & 60.00 & 29.64 & 46.43 \\
& ASFT & 59.20 & 82.40 & 18.98 & 40.81 & 23.28 & 46.52 & 4.17 & 13.33 & 32.81 & 70.00 & 27.69 & 50.61 \\
& TALR & 56.60 & 82.60 & 17.97 & 41.54 & 22.22 & 48.00 & 1.67 & 6.67 & 31.56 & 62.50 & 26.00 & 48.26 \\
& PriFT & 57.75 & 83.20 & 13.28 & 38.24 & 26.85 & 48.44 & 5.42 & 26.67 & 35.62 & 77.50 & 27.78 & 54.81 \\
& SPFT & 64.75 & 87.40 & 20.82 & 40.44 & 28.94 & 51.26 & 7.08 & 16.67 & 43.12 & 82.50 & 32.94 & 55.65 \\
& \textbf{SPFT-SR} & 67.30 & 85.80 & 20.82 & 41.91 & 30.80 & 51.70 & 9.17 & 23.33 & 43.44 & 80.00 & 34.31 & 56.55 \\
\midrule
\multirow{8}{*}{\makecell[c]{Qwen3\\1.7B-Base}}
& Base & 32.07 & 76.00 & 11.35 & 33.82 & 15.85 & 40.74 & 2.92 & 13.33 & 15.62 & 50.00 & 15.56 & 42.78 \\
& SFT & 53.10 & 81.40 & 15.53 & 36.40 & 20.46 & 45.93 & 4.58 & 16.67 & 25.94 & 62.50 & 23.92 & 48.58 \\
& DFT & 54.70 & 75.80 & 18.01 & 31.99 & 19.70 & 40.74 & 3.33 & 16.67 & 29.38 & 55.00 & 25.02 & 44.04 \\
& ASFT & 55.40 & 80.80 & 15.81 & 33.46 & 21.48 & 45.93 & 2.50 & 13.33 & 24.06 & 57.50 & 23.85 & 46.20 \\
& TALR & 53.08 & 80.00 & 16.18 & 37.87 & 21.20 & 45.19 & 3.75 & 13.33 & 30.00 & 67.50 & 24.84 & 48.78 \\
& PriFT & 56.72 & 81.60 & 18.01 & 37.50 & 22.26 & 45.63 & 3.33 & 13.33 & 28.75 & 65.00 & 25.81 & 48.61 \\
& SPFT & 56.15 & 81.00 & 16.64 & 41.54 & 22.67 & 45.63 & 2.92 & 10.00 & 30.94 & 67.50 & 25.86 & 49.13 \\
& \textbf{SPFT-SR} & 57.75 & 80.20 & 19.30 & 35.66 & 23.09 & 45.93 & 7.08 & 20.00 & 30.00 & 62.50 & 27.44 & 48.86 \\
\bottomrule
\end{tabular}}
\end{table*}

\subsection{Majority-Vote (\texttt{Maj@8}) Results for the Main Mathematical Comparison}
\label{app:math_majority_table}

Table~\ref{tab:math_majority} reports the per-benchmark majority-vote accuracy (\texttt{Maj@8}) and its unweighted five-benchmark average. It uses exactly the same eight sampled trajectories per problem as the Avg@8 / Pass@8 results in Table~\ref{tab:math_main}. Across the three main model backbones, SPFT and SPFT-SR achieve the highest average majority-vote accuracy, improving over standard SFT by +8.10 percentage points on Qwen2.5-Math-1.5B (41.67\% vs.\ 33.57\%), +8.94 percentage points on Qwen2.5-Math-7B (51.12\% vs.\ 42.18\%), and +12.78 percentage points on Qwen3-8B-Base (54.94\% vs.\ 42.16\%). The Qwen3-1.7B-Base block is included as an additional model-scale comparison.

\begin{table*}[t]
\centering
\caption{Majority-vote accuracy (Maj@8) across all five mathematical reasoning benchmarks and their unweighted average for the main comparisons and the additional Qwen3-1.7B-Base model-scale comparison.}
\label{tab:math_majority}
\small
\setlength{\tabcolsep}{6.5pt}
\renewcommand{\arraystretch}{1.08}
\resizebox{\textwidth}{!}{\begin{tabular}{llrrrrrr}
\toprule
\textbf{Model} & \textbf{Method} & \textbf{MATH-500} & \textbf{Minerva} & \textbf{Olympiad} & \textbf{AIME24} & \textbf{AMC23} & \textbf{Average} \\
\midrule
\multirow{8}{*}{Qwen2.5-Math-1.5B}
& Base & 56.20 & 6.62 & 31.11 & 13.33 & 27.50 & 26.95 \\
& SFT & 68.00 & 20.59 & 30.07 & 6.67 & 42.50 & 33.57 \\
& DFT & 67.20 & 25.37 & 27.85 & 6.67 & 35.00 & 32.42 \\
& ASFT & 67.60 & 24.26 & 30.07 & 3.33 & 42.50 & 33.55 \\
& TALR & 68.20 & 25.74 & 31.26 & 3.33 & 45.00 & 34.71 \\
& PriFT & 69.20 & 14.34 & 33.63 & 6.67 & 40.00 & 32.77 \\
& SPFT & 74.00 & \textbf{27.57} & 34.81 & 13.33 & 55.00 & 40.94 \\
& \textbf{SPFT-SR} & \textbf{76.00} & 23.90 & \textbf{37.63} & 13.33 & \textbf{57.50} & \textbf{41.67} \\
\midrule
\multirow{8}{*}{Qwen2.5-Math-7B}
& Base & 75.00 & 19.49 & 34.22 & 26.67 & 65.00 & 44.08 \\
& SFT & 76.60 & 33.09 & 41.19 & 10.00 & 50.00 & 42.18 \\
& DFT & 73.00 & 26.47 & 36.44 & 6.67 & 47.50 & 38.02 \\
& ASFT & 77.40 & 25.00 & 41.63 & 6.67 & 65.00 & 43.14 \\
& TALR & 75.20 & 26.84 & 38.52 & 10.00 & 45.00 & 39.11 \\
& PriFT & 79.20 & 25.37 & 38.81 & 20.00 & 67.50 & 46.18 \\
& SPFT & \textbf{81.40} & 29.41 & \textbf{43.11} & \textbf{26.67} & \textbf{75.00} & \textbf{51.12} \\
& \textbf{SPFT-SR} & 77.80 & \textbf{36.40} & \textbf{43.11} & 23.33 & 67.50 & 49.63 \\
\midrule
\multirow{8}{*}{Qwen3-8B-Base}
& Base & 82.60 & \textbf{40.07} & 48.89 & 20.00 & 72.50 & 52.81 \\
& SFT & 76.80 & 23.90 & 39.26 & 13.33 & 57.50 & 42.16 \\
& DFT & 79.00 & 23.53 & 43.11 & 23.33 & 60.00 & 45.79 \\
& ASFT & 81.00 & 24.63 & 46.37 & 23.33 & \textbf{82.50} & 51.57 \\
& TALR & 79.00 & 23.53 & 42.81 & 23.33 & 67.50 & 47.23 \\
& PriFT & \textbf{85.80} & 31.99 & 50.67 & 26.67 & 72.50 & 53.53 \\
& SPFT & 84.80 & 28.68 & 50.81 & \textbf{33.33} & 75.00 & 54.52 \\
& \textbf{SPFT-SR} & 85.20 & 30.51 & \textbf{52.30} & 26.67 & 80.00 & \textbf{54.94} \\
\midrule
\multirow{8}{*}{Qwen3-1.7B-Base}
& Base & 38.60 & 13.24 & 21.33 & 0.00 & 17.50 & 18.13 \\
& SFT & 65.60 & 20.96 & 28.74 & 3.33 & 32.50 & 30.23 \\
& DFT & 61.60 & 20.96 & 25.04 & 3.33 & 30.00 & 28.19 \\
& ASFT & 65.40 & 17.65 & 28.74 & 3.33 & 37.50 & 30.52 \\
& TALR & 65.40 & 20.59 & 28.74 & 3.33 & 45.00 & 32.61 \\
& PriFT & 66.00 & 21.69 & 30.81 & 6.67 & 40.00 & 33.03 \\
& SPFT & 65.20 & 19.12 & 29.33 & 3.33 & \textbf{47.50} & 32.90 \\
& \textbf{SPFT-SR} & \textbf{66.80} & \textbf{25.00} & 30.52 & 10.00 & 42.50 & \textbf{34.96} \\
\bottomrule
\end{tabular}}
\end{table*}

\end{document}